\pdfoutput=1

\documentclass[11pt]{article}

\usepackage[]{emnlp2021}

\usepackage{times}
\usepackage{latexsym}
\usepackage{cleveref}
\usepackage{colortbl}
\usepackage{tabularx,booktabs,multirow}
\usepackage{graphicx}
\usepackage{subfig}
\usepackage[T1]{fontenc}

\usepackage[utf8]{inputenc}

\usepackage{microtype}

\definecolor{airforceblue}{rgb}{0.36, 0.54, 0.66}
\definecolor{ao}{rgb}{0.0, 0.0, 1.0}
\definecolor{darkblue}{rgb}{0.0, 0.0, 0.55}
\definecolor{premise}{RGB}{40, 116, 166}
\definecolor{hypothesis}{RGB}{40, 116, 166}
\definecolor{mask}{RGB}{254, 249, 231}
\definecolor{label}{RGB}{245, 183, 177}

\crefformat{section}{\S#2#1#3} 
\crefformat{subsection}{\S#2#1#3}
\crefformat{subsubsection}{\S#2#1#3}

\title{Trusting RoBERTa over BERT: Insights from CheckListing the Natural Language Inference Task}

\author{Ishan Tarunesh\thanks{\hspace{5pt}Work done when author was a student at IIT Bombay.} \\
  Samsung Korea\\
  \small\texttt{ishantarunesh@gmail.com} \\\And
  Somak Aditya \and Monojit Choudhury \\
  Microsoft Research India, Bengaluru, India \\
  \small\texttt{\{t-soadit,monojitc\}@microsoft.com} \\}

\begin{document}
\maketitle
\begin{abstract}
The recent state-of-the-art natural language understanding (NLU) systems often behave unpredictably, failing on simpler reasoning examples. Despite this, there has been limited focus on quantifying progress towards systems with more predictable behavior. We think that reasoning capability-wise behavioral summary (proposed in \citet{ribeiro-etal-2020-beyond}) is a step towards bridging this gap. We create a \textsc{CheckList} test-suite (184K examples) for the Natural Language Inference (NLI) task, a representative NLU task.
 We benchmark state-of-the-art NLI systems on this test-suite, which reveals fine-grained insights into the reasoning abilities of BERT and RoBERTa. 
 Our analysis further reveals inconsistencies of the models on examples derived from the same template or distinct templates but pertaining to same reasoning capability,  indicating that generalizing the models' behavior through observations made on a CheckList is non-trivial. Through an user-study, we find that users were able to utilize behavioral information to generalize much better for examples predicted from RoBERTa, compared to that of BERT.
\end{abstract}

\section{Introduction}
According to sociologists and the XAI literature \cite{jacovimiller21}, a pre-requisite to \textit{extrinsic} human-AI trust establishment is for users to be able to \textit{anticipate} the model's behavior. In NLP, while we expect pre-trained language models (PTLM) to power agents interacting with humans, often Transformers-based state-of-the-art architectures (BERT \cite{devlin2019bert} and its variants) behave unpredictably showing poor generalization abilities for simpler non-adversarial examples \cite{kaushik2019learning,probinglogical} while achieving state-of-the-art in complex examples requiring composite reasoning \cite{wang2018glue}. This has motivated researchers to re-think \textit{evaluation} methodologies, which is a key component of \textit{extrinsic} trust. Recently, inspired by behavioral testing \cite{beizer1995black}, authors in \cite{ribeiro-etal-2020-beyond} proposed creation of template-based test-suites, called \textsc{CheckList}, that has a broader coverage ranging from the minimal expected functionality to more complicated tests across a range of capabilities. Moving beyond capability-wise probing and cloze-task formulations, the methodology produces a behavioral summary that aggregates different shortcomings of the SOTA models across capabilities in a disentagled manner.  In this work, we hypothesize that the behavioral summary using the \textsc{CheckList} method for a natural language understanding task will help humans form a holistic intuition, that will in turn form the basis of quantifying the \textit{predictability} of models through humans. 

 To test this \textit{central} hypothesis of our work, we choose the Natural Language Inferencing (NLI) task as it tests reasoning capabilities explicitly, and create a \textsc{CheckList} that enables evaluation of whether NLI systems exhibit such reasoning capabilities. In the process, we extend the list of capabilities in \newcite{ribeiro-etal-2020-beyond} to cover more interesting linguistic and logical reasoning phenomena (such as causal, spatial, pragmatic) required in NLI (or similar tasks). We discuss how we come up with templates for such reasoning capabilities. The evaluation results on \textsc{CheckList} test-suite provide a fine-grained disentangled view of a model's capabilities, untangling the effects of different phenomena. However, for models such as BERT \cite{devlin2019bert} and RoBERTa \cite{liu2019roberta}, we discover capability-wise and intra-template inconsistencies. Even though, the average aggregate accuracies tell a clear story, such inconsistencies are found in most systems we evaluated. As a potential resolution, we design a human study and through a simulation experiment \cite{DBLP:journals/corr/abs-1902-00006}, we see how human judgement can be used to quantify model predictability. 

Our contributions are the following. 1) First, we create a template-based test-suite\footnote{We will make the dataset available for public use.} (194 templates, 184k examples) for the NLI task by extending a recently published reasoning taxonomy for NLI \cite{joshi2020taxinli}, and benchmark SOTA NLI systems that reveals new interesting facts about them. 2) We observe inconsistencies in the performance of the models within templates as well as across similar templates (perturbations of the same template), 3) performance inconsistency within templates (across varying lexicon) reveals new biases for BERT, and 4) Through a user study using \textit{simulation} experiments, we provide an indication of human judgement about how inconsistencies affect predictability of models. Particularly, our experiments collectively indicate that RoBERTa is more ``robust" (indeed!) and predictable than BERT. 

\section{Related Work}

The conflicting performance of Transformer-based PTLMs in large natural language understanding benchmarks and targeted phenomena-wise tests have led to a wave of work in probing and attempting to understand these models. Extensive probing tasks have been implemented in order to investigate how and where (within the model) linguistic information has been encoded \cite{tenney2019bert,tenney2019you,hewitt-manning-2019-structural,jawahar-etal-2019-bert,liu2019linguistic,kim-etal-2019-probing}. However, the effectiveness of the leading evaluation methodology, aka probing tasks, have come into question. For example, \newcite{ravichander-etal-2020-systematicity} cautions that BERT may not understand some ``concepts'' even though probing studies may indicate otherwise. 

At a semantic level, and more specifically with respect to the NLI task, inference datasets have been curated focusing on testing a range of reasoning capabilities \cite{poliak-etal-2018-collecting,Richardson2019ProbingNL}. Several work \cite{mccoy-etal-2019-right,kaushik2019learning,glockner-etal-2018-breaking} developed targeted evaluation sets to adversarially challenge these large PTLMs and demonstrated shortcomings. However, these methods rely on the aggregate statistic of accuracy to assess performance, which makes it tricky to pinpoint where exactly the model is failing, 
and how to remedy the issues \cite{wu-etal-2019-errudite}. The recent work by \citet{ribeiro-etal-2020-beyond} takes a different route. Inspired by software testing, authors propose creation of a set of model-agnostic test cases that capture basic expected functionality from a trained system. The revelation that SOTA systems fail on such minimal functionality tests has motivated the community to look at behavioral testing methodologies more closely, through which we can define capabilities and test them individually in a scalable manner. 

Language understanding tasks such as NLI introduces two additional challenges. Understanding requires a set of theoretically well-defined types of reasoning capabilities, as put forward by the theories of semantics and logic \cite{sowa2010role,wittgenstein-1922}. Such types define the necessary capabilities that an NLU (or NLI) system should possess; some of which are missing in the \textsc{CheckList} work \cite{ribeiro-etal-2020-beyond}. A goal of evaluation is also to develop a holistic intuition about model's behavior, and the behavioral summary from \textsc{CheckList} by itself may not be sufficient in achieving such a goal. These central challenges are relevant for all NLU tasks, and constitute the primary focus of our work.

\begin{table*}[!ht]
\setlength\fboxsep{1pt}
\resizebox{\textwidth}{!}{%
\scriptsize
\setlength\tabcolsep{2pt}
\begin{tabular}{p{0.70cm} p{13.5cm} c}
\toprule
\# & \multicolumn{1}{c}{Template \& Examples} & Annotations\\
\arrayrulecolor{black}\midrule

T1 & {\textcolor{premise}P:} \{NAME\} is \{ADJ\}. {\textcolor{hypothesis}H:} \{NAME\} is \{Antonym(ADJ\}. {\colorbox{pink}{\texttt{contradict}}} & \multirow{2}{*}{C (1)}\\
Lex. & {\textcolor{premise}P:} Jim is responsible. {\textcolor{hypothesis}H:} Jim is irresponsible.\\
\arrayrulecolor{black}\midrule
T2 & {\textcolor{premise}P:} \{NAME1\} is \{a/an\} \{PROF\} and \{NAME2\} is too. {\textcolor{hypothesis}H:} \{NAME2\} is \{a/an\} \{PROF\}. {\colorbox{pink}{\texttt{entail}}} &\multirow{2}{*}{E (1)}\\
Synt. & {\textcolor{premise}P:} Kevin is a politician and Steve is too. {\textcolor{hypothesis}H:} Steve is a politician.\\
\arrayrulecolor{black}\midrule
T3 & {\textcolor{premise}P:} \{NAME1\} and \{NAME2\} are from \{CTRY1\} and \{CTRY2\} respectively. {\textcolor{hypothesis}H:} \{NAME1\} is from \{CTRY1\}. {\colorbox{pink}{\texttt{entail}}} &\multirow{2}{*}{E (1)}\\
Bool. & {\textcolor{premise}P:} George and Michael are from Germany and Australia respectively. {\textcolor{hypothesis}H:} George is from Germany.\\
\arrayrulecolor{black}\midrule
T4 & {\textcolor{premise}P:} \{NAME1\} and \{NAME2\} are from \{CTRY1\} and \{CTRY2\} respectively. {\textcolor{hypothesis}H:} \{NAME1\} is from \{CTRY2\}. {\colorbox{pink}{\texttt{contradict}}} &\multirow{2}{*}{C (1)}\\
Bool. & {\textcolor{premise}P:} Helen and Barbara are from Canada and Brazil respectively. {\textcolor{hypothesis}H:} Helen is from Brazil.\\
\arrayrulecolor{black}\midrule
T5 & {\textcolor{premise}P:} \{MALE\_NAME\} and \{FEMALE\_NAME\} are \{friends/collegues/married\}. He is \{a/an\} \{PROF1\} and she is \{a/an\} \{PROF2\}. {\textcolor{hypothesis}H:} \{MALE\_NAME\} is \{a/an\} \{PROF1\}. {\colorbox{pink}{\texttt{entail}}} &\multirow{2}{*}{E (.8)}\\
Coref & {\textcolor{premise}P:} Angelique and Ricardo are collegues. He is a minister and she is a model. {\textcolor{hypothesis}H:} Angelique is a model.\\
\arrayrulecolor{black}\midrule
T6 & {\textcolor{premise}P:} \{CITY1\} is \{N1\} miles from \{CITY2\} and \{N2\} miles from \{CITY3\}. H : \{CITY1\} is \{nearer/farther\} to \{CITY2\} than \{CITY3\}. {\colorbox{pink}{\texttt{entail}}}&\multirow{2}{*}{E (1)}\\
Spatial & {\textcolor{premise}P:} Manchester is 67 miles from Pittsburg and 27 miles from Kansas. {\textcolor{hypothesis}H:} Manchester is nearer to Kansas than Pittsburg.\\
\arrayrulecolor{black}\midrule
T7 & {\textcolor{premise}P:} \{NAME\} has \{EVT1\} followed by \{EVT2\} followed by \{EVT3\}. {\textcolor{hypothesis}H:} \{EVT1/3\} is the \{first/last\} event. {\colorbox{pink}{\texttt{entail}}}&\multirow{2}{*}{E (1)}\\
Temp. & {\textcolor{premise}P:} Barbara has a history class then a mathematics class then a seminar. {\textcolor{hypothesis}H:} The history class is the first event.\\
\arrayrulecolor{black}\midrule
T8 & {\textcolor{premise}P:} \{NAME1\} \{bought/taught/...\} \{OBJ\} to \{NAME2\}. {\textcolor{hypothesis}H:} \{NAME2\} \{sold/learnt/...\} \{OBJ\} from \{NAME1\} {\colorbox{pink}{\texttt{entail}}}&\multirow{2}{*}{E (1)}\\
Causal & {\textcolor{premise}P:} Katherine taught science to Nancy. {\textcolor{hypothesis}H:} Nancy learnt science from Katherine.\\
\arrayrulecolor{black}\midrule
T9 & {\textcolor{premise}P:} \{NAME1\}, \{NAME2\}, ... \{NAMEn\} are the only children of \{NAME0\}. {\textcolor{hypothesis}H:} \{NAME0\} has \{n\} children. {\colorbox{pink}{\texttt{entail}}}&\multirow{2}{*}{E (1)}\\
Num. & {\textcolor{premise}P:} Bill, Patrick, Thomas, Joseph and Scott are the only children of Mark. {\textcolor{hypothesis}H:} Mark has 5 children.\\
\arrayrulecolor{black}\midrule
T10 & {\textcolor{premise}P:} \{NAME1\} is the child of \{NAME0\}, \{NAME2\} is the child of \{NAME0\}, ..., \{NAMEn\} is the child of \{NAME0\}. {\textcolor{hypothesis}H:} \{NAME0\} has \{n1<n\} children. {\colorbox{pink}{\texttt{contradict}}} &\multirow{2}{*}{C (1)}\\
Num. & {\textcolor{premise}P:} Mark is the child of Ruth. Patricia is the child of Ruth. Helen is the child of Ruth. Dorothy is the child of Ruth. Ann is the child of Ruth. {\textcolor{hypothesis}H:} Ruth has 1 child.\\
\arrayrulecolor{black}\midrule
T11 & {\textcolor{premise}P:} \{NAME\} lives in \{CITY\}. {\textcolor{hypothesis}H:} \{NAME\} lives in \{CTRY\}. {\colorbox{pink}{\texttt{entail}}} &\multirow{2}{*}{E (.8)}\\
World & {\textcolor{premise}P:} Patrick lives in Lahore. {\textcolor{hypothesis}H:} Patrick lives in Pakistan\\
\arrayrulecolor{black}\midrule
T12 & {\textcolor{premise}P:} \{NAME\}'s \{RELATION/OBJ\} is \{ADJ\}. {\textcolor{hypothesis}H:} \{NAME\} has a \{RELATION/OBJ\}. {\colorbox{pink}{\texttt{entail}}} &\multirow{2}{*}{E (1)}\\
PreSup& {\textcolor{premise}P:} Sarah's brother is jolly. {\textcolor{hypothesis}H:} Sarah has a brother.\\
\arrayrulecolor{black}\midrule
T13 & {\textcolor{premise}P:} \{NAME\} has stopped \{CONTINUOUS VERB\}. {\textcolor{hypothesis}H:} \{NAME\} used to \{VERB\}. {\colorbox{pink}{\texttt{entail}}} &\multirow{2}{*}{E (1)}\\
PreSup& {\textcolor{premise}P:} Martin has stopped drinking. {\textcolor{hypothesis}H:} Martin used to drink.\\
\arrayrulecolor{black}\midrule
T14 & {\textcolor{premise}P:} \{NAME\} \{PAST VERB\} \{some/few\} of the \{NOUN\}. {\textcolor{hypothesis}H:} \{NAME\} didn't \{VERB\} all of the \{NOUN\}. {\colorbox{pink}{\texttt{entail}}} &\multirow{2}{*}{E (1)} \\
Implic.& {\textcolor{premise}P:} Helen hired some of the teachers. {\textcolor{hypothesis}H:} Helen didn't hire all of the teachers. \\
\arrayrulecolor{black}\midrule
T15 & {\textcolor{premise}P:} \{OBJ1\} and \{OBJ2\} lie on the table. \{NAME\} asked for the \{OBJ1\}. {\textcolor{hypothesis}H:} \{NAME\} did not ask for the \{OBJ2\}. {\colorbox{pink}{\texttt{entail}}} &\multirow{2}{*}{N (.5) E (.5)}\\
Implic.& {\textcolor{premise}P:} Toothpaste and eyeliner lie on the table. Jane asked for the eyeliner. {\textcolor{hypothesis}H:} Jane did not ask for the toothpaste.\\
\arrayrulecolor{black}\midrule
T16 & {\textcolor{premise}P:} \{NAME1\} asked if \{NAME2\} has \{N1\} dollars. \{NAME2\} had \{N2 < N1\} dollars. {\textcolor{hypothesis}H:} \{NAME2\} didn't have \{N1\} dollars. {\colorbox{pink}{\texttt{entail}}} &\multirow{2}{*}{E (1)}\\
Implic.& {\textcolor{premise}P:} Donald asked if Chris had 200 dollars. Chris had 90 dollars. {\textcolor{hypothesis}H:} Chris didn't have 200 dollars. \\
\arrayrulecolor{black}\bottomrule
\end{tabular}
}
\caption{We show a few representative Templates (examples), top human annotated label, and associated confidence (0-1). Full list in Appendix.}
\label{tab:exampletemplates}
\end{table*}

\section{A \textsc{CheckList} for the NLI task}
The \textsc{CheckList} methodology \cite{ribeiro-etal-2020-beyond} assists users in testing NLP models by creating templates for a variety of linguistic capabilities, coupled with \textit{test types} (Minimal Functionality tests (MFTs), Invariance tests (INVs), and Directed Expectation tests (DIRs)) which make the corresponding capability easy to test. These templates can then be used to generated multiple examples using the \textsc{CheckList} tool\footnote{\url{https://github.com/marcotcr/checklist}}.
\begin{table}[!ht]
\resizebox{\columnwidth}{!}{%
\begin{tabular}{@{}lc@{}}
\toprule
\multicolumn{1}{c}{Template} & Expected \\ \midrule
Template: P: \textbf{\textcolor{darkblue}{\big\{NAME\big\}}} is \textbf{\textcolor{darkblue}{\big\{ADJ\big\}}}. H: \textbf{\textcolor{darkblue}{\big\{NAME\big\}}} is \textbf{\textcolor{darkblue}{\big\{Synonym(ADJ)\big\}}} & Entail \\
Example: P: \colorbox{mask}{Alexia} is \colorbox{mask}{happy}. H: \colorbox{mask}{Alexia} is \colorbox{mask}{glad}. & Entail \\ \bottomrule
\end{tabular}%
}
\caption{An example NLI template testing synonyms.}
\label{tab:template1}
\end{table}
 An example \textsc{CheckList} template for NLI task, is shown in Table~\ref{tab:template1}.
Here \textcolor{darkblue}{NAME}, \textcolor{darkblue}{ADJ} are placeholders. Corresponding lexicons are: \textcolor{darkblue}{\{NAME\}} = \{Alexia, John, Mia, ...\}, \textcolor{darkblue}{\{ADJ\}} = \{good, bad, kind, ...\}. \textcolor{darkblue}{Synonym(.)} stands for a synonym of the word (from WordNet). 
The capabilities discussed in \newcite{ribeiro-etal-2020-beyond} are targeted to test the robustness of NLP systems against a minimal set of properties that are necessary yet feasible to check. However in tasks such as NLI, inferencing often requires (one or more) linguistic and logical reasoning capabilities. Our goal is to test the systems against such reasoning types. Even if some reasoning types are deemed not necessary, such tests cumulatively should inform about the systems' abilities in a holistic manner. 
This presents us with two challenges for templated
test-suite generation for the NLI task (or tasks that require different types of reasoning): 1) careful selection of capabilities that represent well-known linguistic and logical taxonomy, and are easily extensible, 2) template creation for such capabilities.

\subsection{Selection of Capabilities}
\label{subsec:taxinli}
According to the linguistics literature \cite{wittgenstein-1922,Jurafsky+Martin:2009a}, deciphering \textit{meaning} from natural language \textit{form} often takes both semantic and pragmatic understanding abilities. From the perspective of Logic (Charles Peirce), there are three pre-dominant forms of reasoning: deductive, inductive, and abductive; that can be employed by an agent to \textit{understand} and \textit{interact}. Other than a few recent work \cite{bhagavatula2020abductive,jeretic-etal-2020-natural}, most of the NLI datasets have widely covered (monotonic) deductive inferences; and lexical, syntactic and semantic understanding abilities. Recently, \citet{joshi2020taxinli} proposed an (extensible) categorization of the reasoning tasks involved in NLI.  This categorization strikes a balance between the high-level categorizations from Language and Logic, while refining the categories and their granularity based on their relevance with respect to current public NLI datasets. Authors define three broad groups of reasoning: {\sc Linguistic}, {\sc Logical} and {\sc Knowledge}, which aligns with our philosophy.  Other categorizations \cite{nie2019adversarial,wang2018glue}, though relevant, are often incomplete as they are tuned towards analyzing errors. 

Here, we introduce the categories briefly and show examples (and templates) in Tab.~\ref{tab:exampletemplates}.
{\sc Linguistic} represents NLI examples where inference process is internal to the provided text; further divided into three categories {\tt lexical}, {\tt syntactic} and {\tt factivity}. 
{\sc Logical} denotes examples where inference may involve processes external to text, and grouped under {\it Connectives}, and {\it Deduction}. {\it Connectives} involve categories such as {\tt negation, boolean, quantifiers, conditionals} and {\tt comparatives}.  \textit{Deduction} involves different types of reasoning such as {\tt relational, spatial, causal, temporal}, and {\tt coreference}. {\sc Knowledge} indicates examples where external ({\tt world}) or commonly assumed ({\tt commonsense}) knowledge is required for inferencing. For detailed definitions, we refer the readers to \citet{joshi2020taxinli}.

\paragraph{Extending TaxiNLI.} We extend the taxonomy by adding back the (pruned) {\tt Numerical} category. We also add a high-level category {\sc Pragmatic}, with two sub-categories {\tt pre-supposition} and {\tt implicatures}. Templates belonging to {\tt factivity} fall under the more general capability pre-supposition. 

\subsection{Template Generation for Reasoning Categories}

We aim to test the minimal expected functionalities along each reasoning type individually. Automatic template creation from public datasets is not straightforward, as examples in public NLI datasets represent multiple capabilities \cite{joshi2020taxinli}.  Even targeted datasets such as Winograd Schema Challenge \cite{levesque2012winograd}  may require careful re-annotation, as examples may require lexical or boolean reasoning. 
 Instead, we resort to manually creating templates and use human annotations to verify the correctness of templated instances. As and when required, we extend the list of basic key placeholders (and corresponding lexicon) provided by the \textsc{CheckList} tool, such as, \textcolor{darkblue}{\big\{PROFESSION\big\}} = \{doctor, actor, politician, $\ldots$\}, \textcolor{darkblue}{\big\{COM ADJ\big\}} = \{smarter, taller\, $\ldots$\}, \textcolor{darkblue}{\big\{CITY\big\}} = \{Paris, New York\, $\ldots$\}\footnote{The complete list is in Appendix.}. We share the list of templates (target phenomena and generated data) with the dataset. Here, we discuss some challenges we face during template creation, and highlight interesting templates (displayed in Tab.~\ref{tab:exampletemplates}).
\\\noindent
\textbf{Linguistic.}~~~
For \underline{Syntactic}, our templates test different types of ellipsis (T2) that require syntactic understanding of the premise and the hypothesis. Paraphrasing is hard to test individually, as most paraphrases in existing-paraphrase corpora \cite{dolan2005automatically,WinNT} are not necessarily entailments, and such paraphrases may involve lexical changes as well. 
\\\noindent
\textbf{Logical.}~~~ 
For \underline{Boolean}, apart from testing logical \textit{and} ($\land$), \textit{or} ($\lor$); we test ordered resolution as well (T3, T4). 
\underline{Quantifier} templates test the understanding of \textit{universal} (all, $\forall$), and \textit{existential} (some, none; $\exists,\neg\exists$) quantification, and the effects of interchanging them. For \underline{Coreference}, we come up with representative templates to test gendered (T5), animate vs. inanimate resolution. 
For \underline{Spatial} templates, we utilize the list of spatial prepositions and adverbs indicating relative positions (near, far, above, below, left, right). We include a set of templates testing cardinal directions (north, east, south and west); and some requiring comparison of distances (T6, requiring both Spatial and Numerical understanding).
\underline{Temporal} templates cover relative occurrences of events using prepositions, such as \textit{before}, \textit{after}, and \textit{until}. Another set of templates test the understanding of time in the day (8AM comes before 2PM), month or year. We add templates for temporally ordered events  (A happened and then B happened, T7) which require the inference of earliest or latest events. 
\underline{Causal} template creation is tricky since often reasoning beyond \textit{form} is required. However, with specific controls placed, we can generate accurate causal premise-hypothesis pairs. We use a set of action-verb pairs which are complementary (e.g. give-take, give-receive, etc.) to describe corresponding actions between 2 entities and an object (T8).  This still results in limited test cases. Explorations into leveraging knowledge graphs such as ConceptNet, ATOMIC \cite{speer-conceptnet,DBLP:conf/aaai/SapBABLRRSC19} to retrieve appropriate causal action phrases could be a way to tackle the issue.
Most \textsc{Logical} category templates implicitly test the \underline{Relational} (deductive reasoning with relations in text) capability, and hence we add three representative templates. Templates falling under \underline{Numerical} test a basic understanding of counting (T9, T10), addition, subtraction and numerical comparison.
\begin{figure*}[!ht]
  \centering
    \subfloat{\includegraphics[width=0.45\textwidth,height=0.24\textheight]{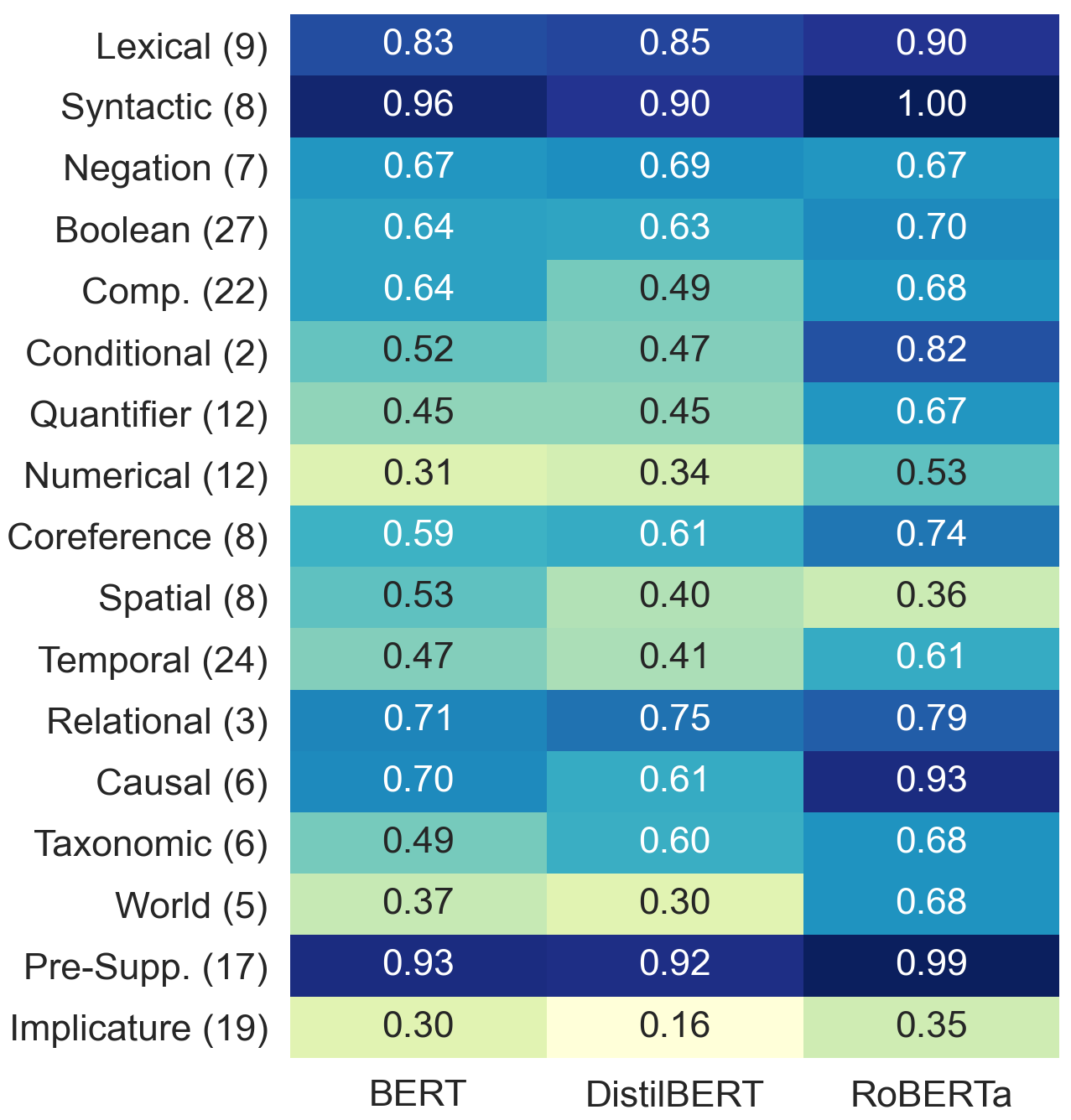}}\hfill
    \subfloat{\includegraphics[width=0.49\textwidth, height=0.24\textheight]{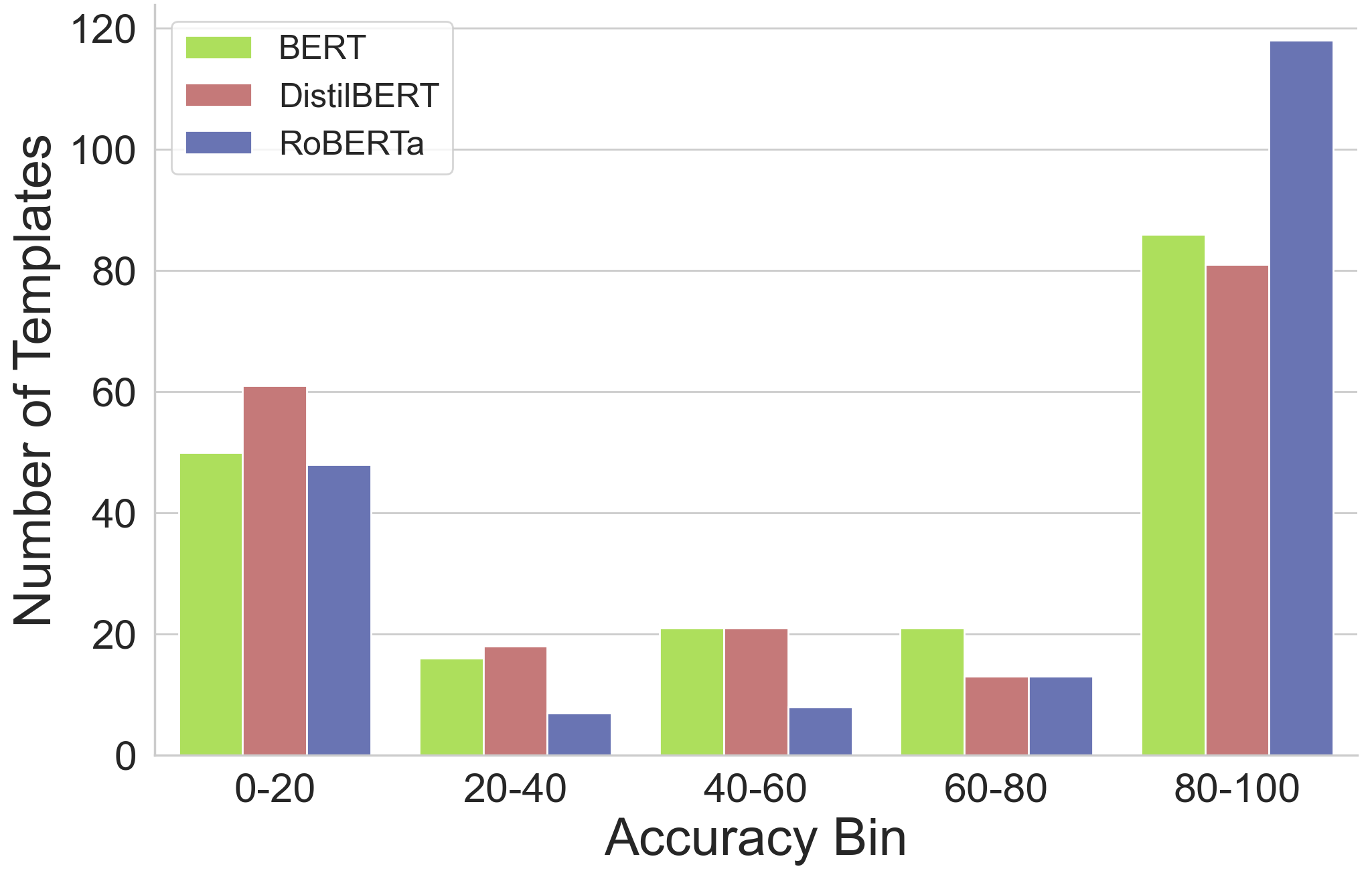}}
    \caption{(a) (Best viewed in color) For each of the 17 reasoning capabilities, we show average accuracy for each model, (b) For 3 models, we show a histogram of number of templates across 5 accuracy bins.
     }
\label{fig:histograms}
\end{figure*}
 \\\noindent
 \textbf{Knowledge. }~~~For \underline{Taxonomic}, we require templates where a taxonomic hierarchy (external to text) is implicitly required to infer the hypothesis. Templates either require to infer A is a type of B (``P: A has some properties. H: A is a type of B.’’;) or utilise that information  (similar to \newcite{NEURIPS2020_e992111e}) (``P: B has property P. H: A has property P.''). We collect set of properties of common flowers\footnote{\url{https://bit.ly/3hu2Psn}}, birds, fishes, and mammals from Wikipedia to generate the templates. The scope of \underline{World} templates is vast. Here, we create templates that specifically tests basic knowledge of geography (city-country pairs, T11), famous personalities (noble prize winners and their contributions\footnote{\url{https://bit.ly/2RkIOdk}}) along with the understanding of some well-known concepts such as \textit{speed} (\textit{speed} decreases when brakes are applied), popularity on social media. \\\noindent
 \textbf{Pragmatic. }
For Pragmatic, we add templates along the lines of \newcite{jeretic-etal-2020-natural}. For \underline{Pre-supposition}, templates test the existence of objects (T12), occurrence of events, aspectual verbs (T13), and quantifiers. \underline{Implicature} templates are constructed by following Grice's cooperative principle (Maxims of quality, quantity, and relevance) \cite{grice1975logic}. Most Implicature templates also require other capabilities such as quantifiers (T14), Boolean (T15), and Numerical (T16).

\section{Benchmarking NLI systems}

 We analyze BERT-base (uncased), DistilBERT-base (uncased) and RoBERTa-large (cased) \cite{devlin2019bert,liu2019roberta} fine-tuned on MultiNLI (referred to as RoBERTa). We also observe the effects of adversarial training using RoBERTa-large fine-tuned on Adversarial NLI dataset (RoBERTa-ANLI; \cite{nie2019adversarial}), a larger model DeBERTa-large \cite{he2020deberta}. For lack of space, we include a short summary of observations from RoBERTa-ANLI and DeBERTa in Appendix. For easy reproduction, we use the MNLI fine-tuned models publicly available from the Huggingface Transformers repository\footnote{\url{https://github.com/huggingface/transformers}}  \cite{Wolf2019HuggingFacesTS}. 


\paragraph{CheckListNLI Dataset.} We created a total of 194 templates spanning all 17 capabilities, discussed in Section  \cref{subsec:taxinli}. For each template (except {\sc Knowledge}), we generate 1000 examples by careful instantiations of the placeholders present in the template.  Since {\sc Knowledge} category involves collecting facts from other sources, we generate 100 examples per template. The dataset will be released upon acceptance. We ask two independent annotators to annotate the NLI label for 5 random examples from each template (970 examples). The average Fleiss' $\kappa$ $0.81$ (same as Cohen's) shows very high inter-annotator agreement.

\subsection{Observations and Analysis}

\begin{table}[]
\small
\centering
\begin{tabular}{cccccc}
\toprule
Dataset & BERT & DistilBERT & RoBERTa \\
\midrule
MNLI-test & 84.5 & 82.2 & 90.2 \\
\midrule
CheckListNLI & 59.4 & 54.6 & 68.2 \\
\bottomrule
\end{tabular}
\caption{Average accuracy on MNLI-test set and CheckListNLI.}
\label{tab:acc_aggregate}
\end{table}

Table~\ref{tab:acc_aggregate} shows the accuracy on MNLI test, and CheckListNLI dataset for all models. Similar to MNLI, RoBERTa clearly outperforms BERT and DistilBERT on \textsc{CheckListNLI}. Further we analyze BERT, DistilBERT and RoBERTa's capability-wise and intra-template performance. 

\paragraph{Capability-wise Performance.} The capability-wise average accuracy of the models are shown in Figure \ref{fig:histograms}(a). We observe that all models perform well on Lexical, Syntactic and Presupposition capabilities. Within the Logical category, the results are comparably poor and inconsistent across both the capabilities and model dimension. The same holds for the Knowledge categories and Implicature templates.
For further analysis, we mark a template as \textit{passed} if the model's accuracy is above $80\%$, as \textit{unsure} if the accuracy is in middle bins ($20$-$80\%$), and \textit{failed} if less than $20\%$. For \underline{negation}, all models fail on template containing ``but not'' (P: Janet, but not Stephen, is a dancer. H: Stephen is a dancer.) For \underline{boolean}, we observe that BERT and DistilBERT are \textit{unsure} on ordered resolution; whereas RoBERTa is biased towards entailment label (\textit{unsure} for contradiction). Moreover for RoBERTa, the bias shifts towards contradiction with the addition of ``not'' in the hypothesis even if the correct label is entailment (P: Margaret and Robert are from America and Russia respectively. H: Margaret is not from Russia.). For \underline{comparative}, all models fail on template with insufficient information (P: Philip is more handsome than Frances. Philip is more handsome than Kevin. H: Kevin is more handsome than Frances.). BERT and DistilBERT are \textit{unsure} on reasoning about hypothesis in the presence of superlative adjective in the premise (P: Among Emily, Daniel, and Joseph, the bravest is Daniel. H: Emily is braver than Daniel.) DistilBERT fails on a template (P: John is taller than Mia. H: Mia is taller than John.) when the placeholders are reversed.  Such perturbations are more common for \underline{Quantifier}. BERT and DistilBERT fail when ``all'' is replaced by ``some'' in the hypothesis. 
 Within \underline{numerical}, RoBERTa seems better at counting, however it fails when the hypothesis refers to an incorrect count. BERT and DistilBERT models seems \textit{unsure} for all counting-related templates. All models struggle with templates related to addition and subtraction (often showing label bias).  Under \underline{Spatial}, all models struggle with cardinal directions and spatial relations (left, right). Interestingly, RoBERTa fails at spatial distance comparisons (while being able to compare number of coins under Numeric). Within \underline{Temporal}, all models are unable to compare year of birth and time of the day. A surprising observation is that BERT is sensitive to the lexical substitution of ``before'' (``after'') with  ``earlier than'' (``later than''). RoBERTa is able to accurately reason ``A happened before/after B'' over two to three events, whereas BERT and DistilBERT results are \textit{unsure}. All models fail to detect the ``first'' or the ``last'' event in a sequence. Within \underline{coreference} RoBERTa is able to accurately resolve male and female names. Lastly, for \underline{Causal} templates, RoBERTa seems more accurate than both BERT and DistilBERT. For \underline{knowledge} templates, all models consistently suffer. On probing \underline{implicature} templates, we observe that models vary between logical (P: Silverware and plate lie on the table. Barbara asked for the plate. H: Barbara also asked for the silverware.; predicting neutral over contradiction) and implicative (P: Some of the balls are purple in colour. H: All of the balls are purple in colour.; predicting contradiction over neutral), depending on the template.

\paragraph{Intra-Template Performance.} In Figure \ref{fig:histograms}(b), we plot the histogram of number of templates across different accuracy bins. The middle bins (20-80) indicate that models are \textit{unsure} on the tested phenomena. This happens most often for BERT (58/194) and DistilBERT (52/194) compared to RoBERTa (28/194). We analyzed the bias of BERT across the vocabulary of placeholders. We examine the template P: \{NAME1\}, but not \{NAME2\}, is a \{PROFESSION\}. H: \{NAME2\} is a \{PROFESSION\}. Template accuracy highly varies when profession is fixed to engineer ($0\%$), dancer ($9\%$) vs. doctor ($80\%$) and professor ($92\%$). Compared to professions, we see only limited variations, when names are restricted to male vs. female names. Similar effects are seen for adjectives in P : \{NAME1\} is \{ADJ\}. \{NAME2\} is \{COM ADJ\}. H : \{NAME2\} is \{COM ADJ\} than \{NAME1\}. Template accuracy varies when adjectives fixed to bigger ($0\%$), sweeter ($16\%$), vs. creepier ($100\%$).

\begin{figure}[!ht]
    \centering
    \includegraphics[width=\columnwidth, height=0.17\textheight]{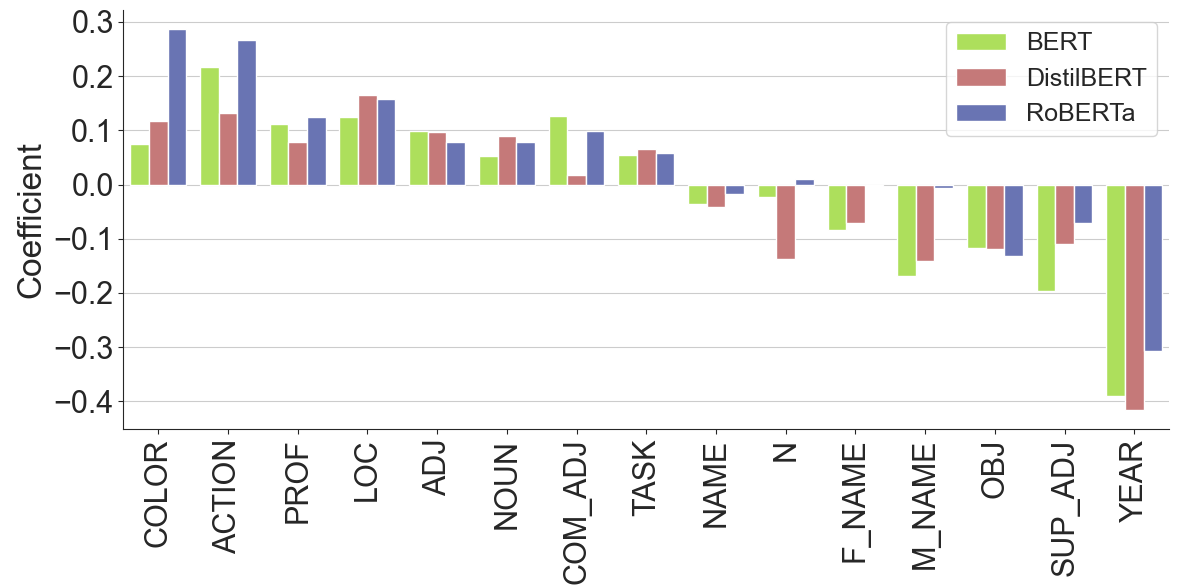}
    \caption{Figure showing the regression coefficients for common placeholders for all three models.}
    \label{fig:placeholders}
\end{figure}

To dive deeper, we first analyze the effect of placeholders on template accuracy using Linear Regression as a surrogate model for feature importance. The feature vector is a one-hot representation created using the concatenation of placeholders, top 20 other words (using BOW) and template label. 
We show the coefficients for placeholders in Fig. \ref{fig:placeholders} (rare placeholders are omitted). 
  Placeholders \textcolor{darkblue}{COLOR} and \textcolor{darkblue}{ACTION} have high positive coefficients as they co-occur with quantifier, syntactic, and pre-supposition templates where the models perform well. Similarly, high negative coefficient for \textcolor{darkblue}{YEAR} is due to models being unable to compare year of birth. Placeholders \textcolor{darkblue}{MALE NAME} and \textcolor{darkblue}{FEMALE NAME} turns out to be interesting having negative coefficient for BERT and DistilBERT, and near-zero coefficient for RoBERTa. This is intuitive as RoBERTa is better at resolving gendered coreferences. Interestingly, \textcolor{darkblue}{NAME} is more negative in BERT, showing the hidden effect of how varying names can affect BERT more than RoBERTa. Similarly, models perform decently on templates involving comparative (\textcolor{darkblue}{COM ADJ}) while struggling in templates involving superlatives (\textcolor{darkblue}{SUP ADJ}). 
  The behavioral analysis gives us an indication that RoBERTa-large may indeed be more robust and accurate, but inter-template inconsistencies begs for further exploration. 

\section{Consulting Humans to Quantify Model Inconsistency}
\label{sec:human}

The inconsistencies observed for both BERT and RoBERTa begs the question as to how to quantify the progress towards models with increased predictable behavior\footnote{Training machine learning models to predict behavior adds more confounders that can affect the analysis.}. Inspired by the recent XAI literature \cite{gilpin2018explaining,lipton2018mythos,DBLP:journals/corr/abs-1902-00006}, we design a human study where humans are shown different types of post-hoc behavioral information and asked to predict system's behavior on new examples.
This premise presents us with many dimensions of control, which may affect the outcome of the study: 1) the level of abstraction of behavioral information, 2) example-based (local) vs. global summary, 3) interface design (or presentation of such information), 4) baseline explanation, 5) choice of test examples, 6) task (verification, simulation or counterfactual), 7) choice of participants. 

\begin{figure}[!ht]
    \centering
    \includegraphics[width=\columnwidth, height=0.3\textheight]{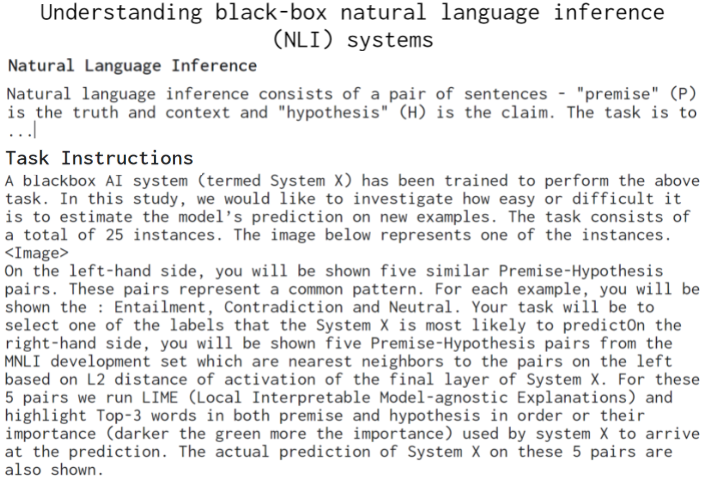}
    \caption{Landing Page Instructions to participants. Study in \url{https://bit.ly/3huowZG}}
    \label{fig:taskif}
\end{figure}

\paragraph{Local Explanations and Interface Design.} From multiple pilot studies using global behavioral summary (such as Template-wise accuracy scores), and relevant template-level local summaries \footnote{Detailed in Appendix. Current study (Stage B1): \url{checklist-nli.herokuapp.com/A}}, we observe that global template-wise accuracies have large cognitive load for participants, and template-wise aggregate accuracies are hard to comprehend (and predict) without the knowledge of the lexicon that the keywords represent (owing to intra-template inconsistencies). Hence, in this study, we fixate on providing local example-based explanations. For the explanations interface design, we follow \citet{DBLP:journals/corr/abs-1902-00006}. Through various pilot studies, authors in \citet{DBLP:journals/corr/abs-1902-00006} found that the use of three intuitive boxes -- namely, the input box, the explanation box, and the question box; makes the information easier to follow. We observe that this vastly improves relative to our earlier pilot studies based on Google Forms-based questionnaire. 

\paragraph{Test Example Selection and Baseline.}  To choose test examples, we choose 25 test templates from CheckListNLI carefully by balancing templates representing different accuracy buckets, and ensuring inclusion of multiple capability-templates. For each template, we show 5 random test examples. As a baseline, we consider example-based LIME \cite{ribeiro2016should} explanations. For each test example, five nearest neighbor examples are chosen, where the top three attended words (from Premise and hypothesis) are highlighted using LIME output. The variations are created by varying where the nearest neighbors come from. In Stage 1, we choose the nearest neighbors from the MultiNLI validation set and Stage 2 includes nearest neighbors from CheckListNLI dataset respectively (barring the corresponding test template examples). To calculate nearest neighbors, we use the underlying system's (BERT/RoBERTa) final hidden-layer embeddings (corresp. to \texttt{[CLS]} token) and calculate cosine distances. 

\paragraph{Task, Metrics and Participants.} In this study, we restrict ourselves to \textit{simulation} questions, where the participants are asked to simulate (\textit{anticipate}) an underlying blackbox system's (instructions in Fig.~\ref{fig:taskif}) prediction given explanation and the input. Since Transformers often show different types of bias and its not known whether they follow human reasoning, we track two metrics: 1) prediction accuracy, 2) mutual agreement score. Consuming such examples and being able to generalize based on explanations on nearest neighbors require a certain amount of analytical reasoning skills. Hence, instead of going to crowd-sourced platforms, we choose a total of 10 International Linguistics Olympiad participants\footnote{\url{https://www.ioling.org/}}. Such participants are trained on analytical text-based puzzle solving, but not trained in formal Linguistics or Logic.


\subsection{Findings and Observations}
\begin{figure}[!ht]
\centering
    \includegraphics[width=\columnwidth]{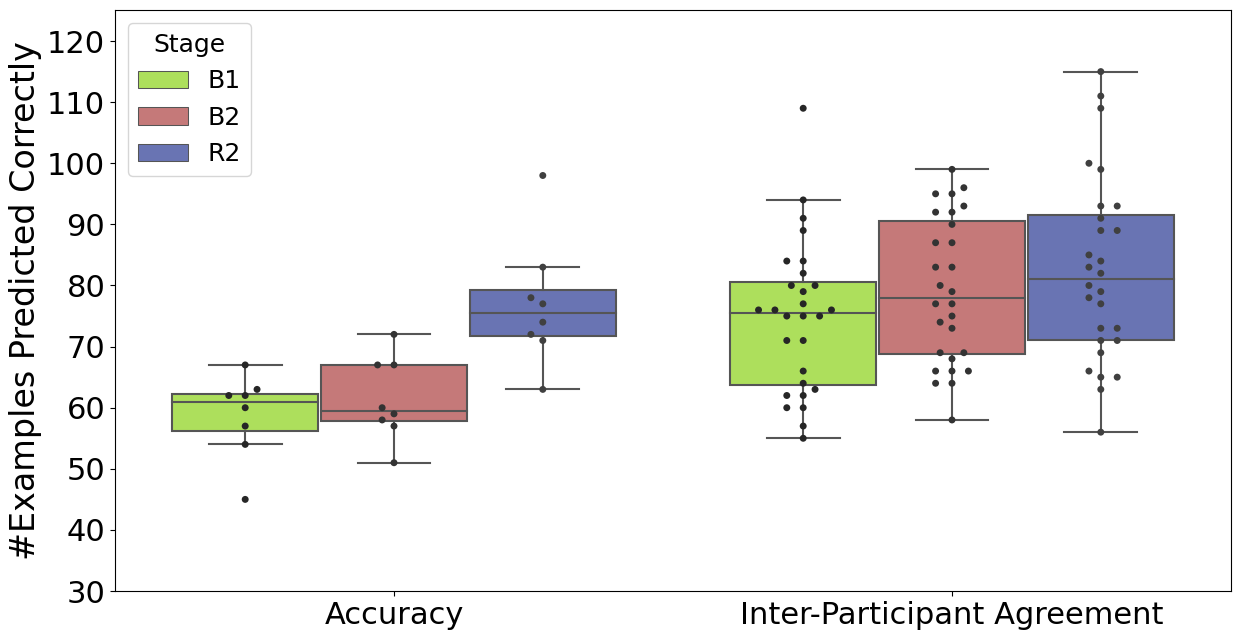}
    \caption{Average Accuracy and Mutual pairwise agreement for participants (out of 125).}
    \label{fig:study2}
\end{figure}
In Figure \ref{fig:study2}, we report results from Stages 1 and 2 for BERT (referred to as B1, B2) and Stage 2 for RoBERTa (R2). The average prediction accuracy of the participants increase from $58.7\pm6.8$ (out of 125) to $61.4\pm6.8$  from Stage B1 to B2. Similarly,  average mutual agreement among participants increases from $74.75$ to $79.8$, showing that the nearest neighbors from CheckListNLI (representing disentangled phenomena) enable better explanation. Upon interviewing participants, some mentioned their responses in stage B1 was often ``random''. This is intuitive as MNLI examples are often quite complex with long sentences. So, we repeat only stage 2 for RoBERTa (R2). For R2, we clearly see a $12.5\%$ improvement in prediction accuracy with a $2.62\%$ improvement in agreement score. This indicates that even though inconsistencies exists for both models, participants were able to anticipate RoBERTa's behavior better.

We also recorded responses about the different stages of study from participants. Most participants found that nearest neighbor examples were most relevant for B2 (5 out of 8) and R2 (4 out of 8). The LIME-based highlights and predicted labels were both useful in B2 and R2. A participant commented that "The highlighted words and predictions were very helpful. Relied completely on both". Interestingly, the participants were only told that the systems are different for B2 and R2, without revealing any further detail. But, most people ($60\%$) found that its easier to predict the system's behavior for R2. Participants mentioned that task interface was easy to navigate and they had no difficulty understanding the instructions.

\section{Conclusion}
Following the recent XAI literature \cite{jacovimiller21}, we aim to quantify progress towards more predictable natural language understanding models (especially PTLMs). To this end, we select the NLI task which requires reasoning, and conduct detailed behavioral analysis of state-of-the-art NLI systems. We adapt and extend a recently proposed taxonomy for NLI (\textsc{TaxiNLI}). Through a templated test-suite (194 templates, 17 reasoning types), we observe that for both BERT and RoBERTa, model inconsistencies can be found both across templates (i.e., logical, lexical perturbation of templates) and within templates while only varying lexicon. Furthermore, we design a human study where we ask humans to predict model behavior given behavioral information about NLI systems. A $12.5\%$ increase in human prediction accuracy for RoBERTa over BERT provides an indication that, despite fine-grained inconsistencies, RoBERTa is more predictable than BERT.
Our work shows how behavioral information may help quantify progress towards systems with more predictable (therefore trusted; \cite{jacovimiller21}) behavior. 

\section*{Ethics Statement}
As our work involves human study of the behavior of a blackbox NLI system, we took an Internal Review Board (IRB) approval from our organization for the study and asked for all necessary consent. A consent form was shared with all participants, and all participants formally agreed to participate by electronically signing the form. To the best of our knowledge we ensured that the study did not expose them to any possible harmful content (in text, images or other forms). We also did not collect or share any personally identifiable information.

\section*{Acknowledgement}
We would like to thank Pratik Joshi for contribution to coreference templates. We would also like to thank Sebastian Santy, Saujas Vadguru and Aalok Sathe for attempting and providing useful insights during human study pilots.

\appendix

\section{Benchmarking: Additional Results}

\begin{figure*}[!ht]
  \centering
    \subfloat{\includegraphics[width=0.49\textwidth]{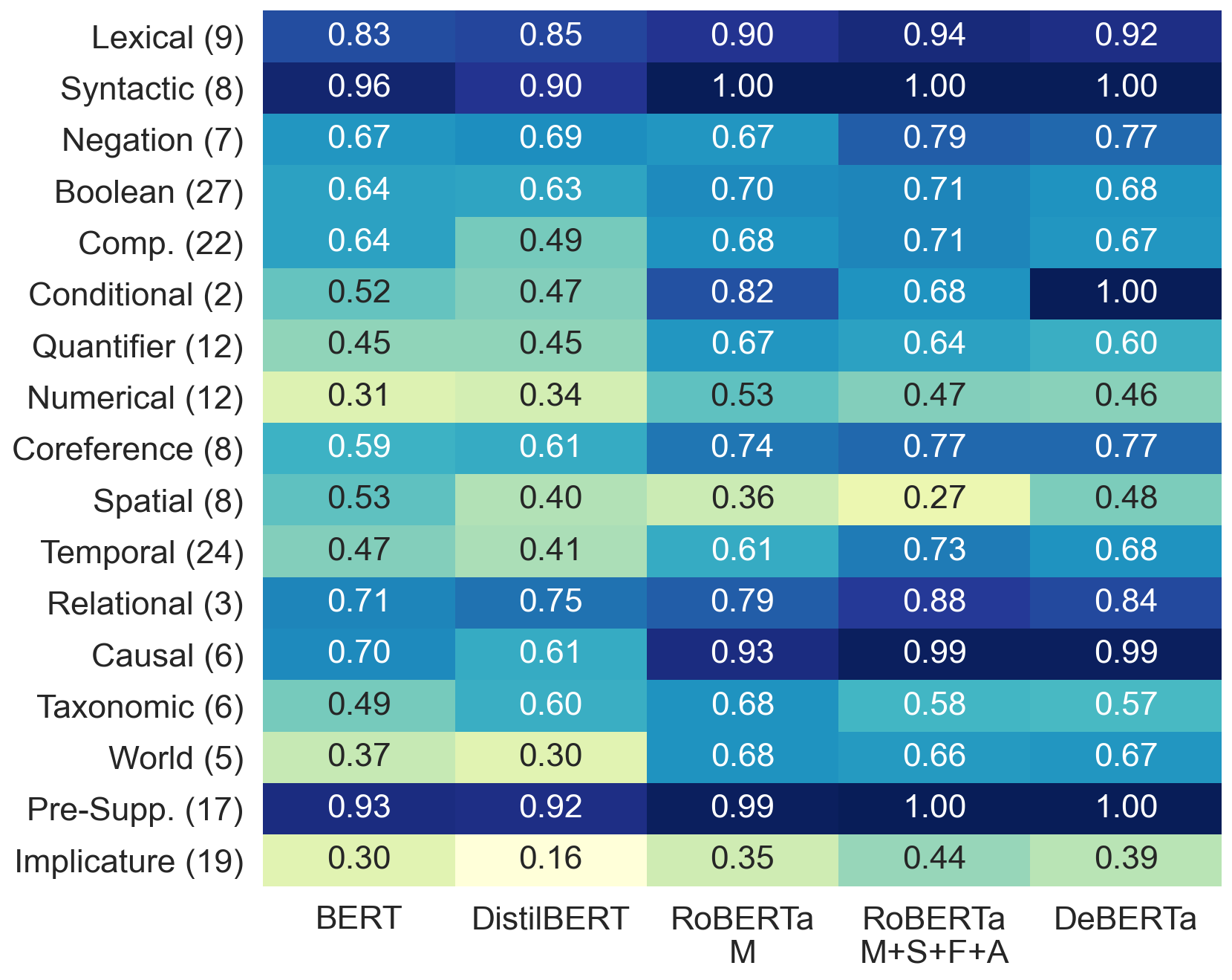}}\hfill
    \subfloat{\includegraphics[width=0.49\textwidth, height=0.24\textheight]{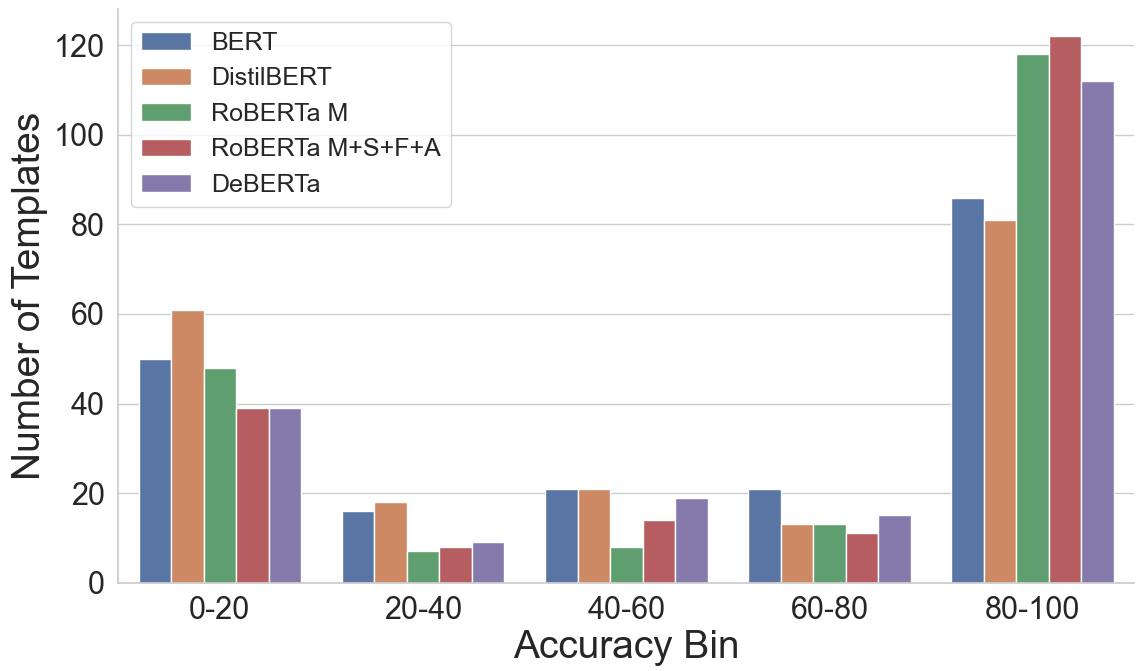}}
    \caption{(a) (Best viewed in color) For each of the 17 reasoning capabilities, we show average accuracy for each model (darker the color, higher the accuracy), (b) We show test-suite wide model accuracies divided into 5 bins.
     }
\label{fig:app:histograms}
\end{figure*}

\begin{table}[!ht]
\centering
\resizebox{\columnwidth}{!}{%
\begin{tabular}{cccccc}
\toprule
Dataset & BERT & DistilBERT & RoBERTa & RoBERTa & DeBERTa \\
& & & MNLI & (M+S+F+A) & \\
\midrule
MNLI-test & 84.5 & 82.2 & 90.2 & - & \textbf{91.3/91.1} \\
\midrule
CheckList & 59.4 & 54.6 & 68.2 & \textbf{71.1} & 69.9 \\
\bottomrule
\end{tabular}%
}
\caption{Average Accuracies of all systems}
\label{tab:app:accuracy_all}
\end{table}
In Tab.~\ref{tab:app:accuracy_all}, we show accuracies of 5 state-of-the-art NLI systems. We observe the effect of adversarial training with more data using the RoBERTa-large trained on Adversarial NLI dataset \cite{nie2019adversarial} (primarily the round 3 model trained on MNLI, SNLI, Fever and ANLI), and using a larger model DeBERTa-large \cite{he2020deberta} trained on Multi-NLI dataset which is at this point leader in the GLUE leaderboard. Interestingly, our capability-wise analysis in Figure.~\ref{fig:app:histograms}(a) shows that DeBERTa does only marginally better than RoBERTa. It still suffers in spatial, numerical, knowledge and implicature templates. In many cases, DeBERTa shows similar intra-template inconsistencies (Fig.~\ref{fig:app:histograms}(b)). RoBERTa-ANLI (M+S+F+A) model, only provides a $2.9\%$ accuracy improvement on CheckListNLI, showing the test-suite is quite hard. Similar to DeBERTa, it suffers in spatial, Numerical, knowledge, implicature and some more logical categories. However, we see from Fig.~\ref{fig:app:histograms}(b), that the intra-template inconsistencies decrease even more for RoBERTa-ANLI. 

\subsection{Benchmarking: Intra-Template Variations}
\begin{table}[!ht]
\centering
\begin{tabular}{ccc}
\toprule
\{PROFESSION\} & Male & Female \\
\midrule
engineer       & 0.00    & 0.01      \\
poet           & 0.00    & 0.04      \\
entrepreneur   & 0.01    & 0.01      \\
politician     & 0.02    & 0.35     \\
writer         & 0.03    & 0.07      \\
banker         & 0.06    & 0.08      \\
dancer         & 0.9    & 0.12     \\
actor          & 0.10   & 0.26     \\
painter        & 0.10   & 0.33     \\
accountant     & 0.12   & 0.35     \\
businessman    & 0.16   & 0.32     \\
author         & 0.33   & 0.45     \\
singer         & 0.49   & 0.77     \\
teacher        & 0.62   & 0.63     \\
doctor         & 0.80   & 0.77     \\
professor      & 0.92   & 0.83    \\
\bottomrule
\end{tabular}%
\caption{Variation in accuracy in BERT for different values of professions and gender}
\label{tab:app:lexical_variation1}
\end{table}

\begin{table}[!ht]
\centering
\begin{tabular}{ccc}
\toprule
\{ADJ\} & Male & Female \\
\midrule
big     & 0.00    & 0.03      \\
sweet   & 0.16   & 0.11     \\
smart   & 0.25   & 0.20     \\
weird   & 0.38   & 0.40     \\
strong  & 0.41   & 0.40     \\
tough   & 0.47   & 0.48     \\
old     & 0.56   & 0.55     \\
tall    & 0.56   & 0.61     \\
tiny    & 0.99   & 0.99     \\
creepy  & 1.00  & 1.00   \\
\bottomrule
\end{tabular}
\caption{Variation in accuracy of BERT for different values of adjective and gender}
\label{tab:app:lexical_variation2}
\end{table}

We expand on our analysis of variation within a template arising due to lexicons by discussing two example templates.  

\paragraph{Case Study 1 on PROFESSION.} 
The expected label for the template {\tt P: \{NAME1\}, but not \{NAME2\}, is a \{PROFESSION\}.} {\tt H: \{NAME2\} is a \{PROFESSION\}. } is ``contradiction''.
We create examples from this template by varying on two dimensions, 1) first we vary the gender of \textcolor{darkblue}{NAME2} (we keep \textcolor{darkblue}{NAME1} to be same gender as \textcolor{darkblue}{NAME2} to avoid any confusion), and then 2) use different values of the \textcolor{darkblue}{PROFESSION} lexicon, and then vary only names. For each combination of the above two we sample 100 examples and report BERT's accuracy in Table \ref{tab:app:lexical_variation1}. As mentioned, accuracy varies strongly with the change in lexicon for \textcolor{darkblue}{PROFESSION}. Template accuracy varies from low when profession is set to engineer ($0\%$), poet ($0\%$) to high when profession is set to doctor ($0.82\%$), professor ($0.92\%$).

\paragraph{Case Study 2 on Adjectives.} 
Next, we select another template: {\tt P: \{NAME1\} is \{ADJ\}. \{NAME2\} is \{COM ADJ\}.}{\tt H: \{NAME2\} is \{COM ADJ\} than \{NAME1\}.} For this template, expected label is entailment.
Similar to before, we generate examples conditioned on \textcolor{darkblue}{ADJ} and gender. Table \ref{tab:app:lexical_variation2} shows the accuracy for different combinations of gender and \textcolor{darkblue}{ADJ}. Compared to the previous template, we notice that the accuracies are almost unchanged across the gender dimension. This means that compared to \textcolor{darkblue}{ADJ}, BERT shows bias towards certain professions \textcolor{darkblue}{PROFESSION} conditioned on gender. Here also template accuracy varies from low when adjectives are bigger ($0\%$), sweeter ($0.16\%$) to  high when adjectives are tiny ($0.99\%$), creepier ($1.00\%$).

\section{Benchmarking: Detailed Observations from Template Perturbations}
We provide a list of interesting templates in Table \ref{tab:app:interestingtemplates}, from where we can glean more fine-grained observations about the underlying systems' behavior. \\
{\tt \underline{T2}:} All models fails on simple boolean template for testing ``or''.\\
{\tt \underline{T13,T14}:} BERT and DistilBERT are {\it unsure} on ordered resolution. RoBERTa is able to predict accurately when the label is entailment but {\it unsure} when the label is contradiction. The observation remains consistent for chain of length 2, 3 and 4. \\
{\tt \underline{T21,T22}:} We modify {\tt T13,T14} by introducing ``not'' in the hypothesis and observe a label shift towards contradiction for all models (even RoBERTa which was accurate for entailment templates).\\
{\tt \underline{T45,T46}:} We look at both these templates together and observe that BERT and DistilBERT are biased towards entailment label while not understanding gendered names. RoBERTa on the other hand performs fairly accurately on both templates. \\
{\tt \underline{T63}:} DistilBERT fails on this very basic comparative template where the arguments are swapped.\\
{\tt \underline{T68,T71}:} The information within the premise is not sufficient to arrive at hypothesis. All models struggle with templates such as this.\\
{\tt \underline{T76,T77}:} This template requires comparative and syntactic understanding. RoBERTa performs accurately on both these template whereas BERT and DistilBERT are {\it unsure}. On further analysis we observe that BERT has lexical bias for the placeholder \textcolor{darkblue}{ADJ}. \\
{\tt \underline{T80,T81}:} All models fail on template related to 2D directions.\\
{\tt \underline{T88,T89,T92,T93}:} All models are {\it unsure} or biased on this set of templates. Since RoBERTa is able to perform numerical comparison we would expect it to compare year but that is not observed. An interesting observation is BERT being sensitive to the lexical substitution ``before'' (or ``after'') to ``earlier than`` (or ``later than'').\\
{\tt \underline{T98,T99}:} RoBERTa is able to correctly reason out the relative ordering to events whereas BERT and DistilBERT are {\it unsure}\\
{\tt \underline{T116,T117}:} Compared to BERT and DistilBERT, RoBERTa is better at understanding causal-verb pairs.\\
{\tt \underline{T122}:} A very simple quantifier template on which both BERT and DistilBERT fail.\\
{\tt \underline{T171,T172}:} These two are exactly the same template with different label depending on ``logical'' vs ``implicative'' reasoning. BERT and DistilBERT are {\it unsure} whereas RoBERTa is ``logical'' .\\
{\tt \underline{T128,T127}:} Another pair of exactly same template but this time all the three models predict contradiction which is ``implicative'' reasoning.\\

\section{Summary of Pilot Studies}

We carry out series of pilot studies varying user interface, explanation granularity and type of behavioral information. We choose 4 (final-year) undergraduate Computer Science students, who have some experience in NLP through projects, but not trained on Linguistics and Logic.

\subsection{Phase 1: Global Explanation, Template-wise Accuracy Prediction}

\begin{table}[!ht]
\centering
\scriptsize
\begin{tabular}{ccccccc}
\toprule
\multicolumn{1}{c}{Stage} & 1 & 2 & 3 & 1 & 2 & 3\\
\arrayrulecolor{black}\midrule
 & \multicolumn{3}{c}{Average Score} & \multicolumn{3}{c}{Argmax Bin Accuracy}\\
\arrayrulecolor{black!30}\midrule
P1 & 1.4 & 1.08 & 1 & 0.28 & 0.24 & 0.16 \\
\arrayrulecolor{black!30}\midrule
P2 & 0.84 & 0.96 & 0.88 & 0.12 & 0.12 & 0.12\\
\arrayrulecolor{black!30}\midrule
P3 & 1.16 & 1.68 & 1.64& 0.24 & 0.40 & 0.40 \\
\arrayrulecolor{black!30}\midrule
P4 & 1.36 & 1.4 & 1.36 & 0.32 & 0.28 & 0.24 \\
\arrayrulecolor{black}\midrule
\multicolumn{1}{l}{} & \multicolumn{3}{c}{Pearson Correlation} & \multicolumn{3}{c}{Argmax Bin Consensus}\\
\arrayrulecolor{black!30}\midrule
P1 \& P2 & 0.41 & 0.14 & 0.29 & 0.52 & 0.52 & 0.60 \\
\arrayrulecolor{black!30}\midrule
P3 \& P4 & 0.02 & 0.60 & 0.62 & 0.32 & 0.72 & 0.60 \\
\arrayrulecolor{black}\bottomrule
\end{tabular}%
\caption{Phase 1 Pilot Study Results using Global observation summaries. Average score (out of 5) is the average bid on the right accuracy bin, across 25 test templates. Pearson Correlation is calculated between the bid put in the correct bin by the two participants. ``Argmax bin'' indicates metrics where we consider where the participants have placed their maximum bid, ignoring the bid  value.  Consensus is the number of times both participants placed their highest bid in the same bin (not necessarily the correct one).}
\label{tab:consensus_phase1}
\end{table}
The first phase of our pilot studies involved a Google Form-based questionnaire as an interface\footnote{\url{https://forms.gle/kzqB2Luqub97mKQd6}}. We envisioned a training stage, where the participants will be shown behavioral summary of a blackbox NLI system; and a testing stage where participants will be asked to place bids (out of 5) on the most accurate bin(s) for unseen templates. First, we chose these 25 test templates, similarly in the final study. Then, we design the three variants (or stages) by varying levels of behavioral information (abstract to fine-grained). In \underline{Stage 1}, we only provided the NLI label-wise accuracies for the models. In \underline{Stage 2}, we provided Template wise accuracies for all remaining templates from the test-suite (at that point 100). Then, on \underline{Stage 3}, we carefully chose 30 related templates and added short textual description for each. Since, the behavioral information was at a global level, the participants found it very hard to successfully utilize the information for each specific template. Moreover, the underlying system BERT was severely inconsistent even within templates. The combined effects of these factors resulted in particpants' predictions to be near random and the performance hardly improved over the first stage (shown in Table.~\ref{tab:consensus_phase1}). However, in some cases, the relevant fine-grained behavioral information seemed to improve agreement score among participants.

\subsection{Phase 2: Local Explanation, Template-level vs. Example-wise Questions}

\begin{table}[!ht]
\centering
\resizebox{\columnwidth}{!}{%
\begin{tabular}{@{}llll@{}}
\toprule
 & \begin{tabular}[c]{@{}l@{}}Template-Based \\ Explanations\end{tabular} & \begin{tabular}[c]{@{}l@{}}Example-based\\ Explanations\end{tabular} & \begin{tabular}[c]{@{}l@{}}Example-based\\ Q+E\end{tabular} \\ \midrule
Accuracy & 14.5 & 17.0 & 22.0 \\
Agreement & 15  & 19.0 & 20.0 \\ \bottomrule
\end{tabular}%
}
\caption{Pilot Study Phase 2 (Based on Website Interface): Average accuracy and agreement scores for local question-wise explanations.  For first two columns, we stick to template-based questions, where participants are required to predict the correct accuracy bin. For the last, we move to example-based questions and explanations. All scores are out of 25.}
\label{tab:app:consensus_phase2}
\end{table}

Using the lessons from the first pilot study, we moved on to local explanations. Using a web-based user interface designed according to \newcite{DBLP:journals/corr/abs-1902-00006}, we carried out pilot studies for overall three variations: 1) template-based questions and explanations, 2) template-based questions and example-based explanations, and 3) example-based questions and explanations. We proceed with same way of selecting 25 test templates and then Template-based questions are formulated in the similar manner as pilot study 1. In each page (for a Test Template), the user is shown a template (and an example), and asked to predict the most accurate bin for the underlying System X out of 5 accuracy bins.  For template-based explanations, we manually chose three useful templates and highlighted a total of 5 examples with LIME output (top 3 words of the premise and the hypothesis). For example-based explanations, we used nearest neighbor examples from CheckListNLI dataset to show 5 nearest examples (again highlighted with LIME output). Lastly, for example-based questions, we removed all information about templates (from instructions and pages). On the questions box, we instead ask participants to predict the entailment labels for 5 random examples from the chosen test template. In the left, we show nearest neighbor example-based explanations, similar to the final study. We clearly see from the pilot study (Tab.~\ref{tab:app:consensus_phase2}), that accuracies gradually improved as we moved towards both example-based questions and explanations.

\begin{table*}[!ht]
\scriptsize
\setlength\fboxsep{1pt}
\resizebox{\textwidth}{!}{
\begin{tabular}{p{0.3cm} p{12.5cm} ccc}
\toprule
\# & \multicolumn{1}{c}{\multirow{2}{*}{Template}} & \multicolumn{3}{c}{Performance (Accuracy \%)} \\
\arrayrulecolor{black!30} \cline{3-5}\noalign{\smallskip}
& & \multicolumn{1}{c}{\tiny BERT} & \multicolumn{1}{c}{\tiny DistilBERT} & \multicolumn{1}{c}{\tiny RoBERTa} \\ 
\arrayrulecolor{black}\midrule
T2 & {\textcolor{premise}P:} \{NAME1\} or \{NAME2\} is \{ADJ\}. {\textcolor{hypothesis}H:} \{NAME1/2\} is \{ADJ\}. {\colorbox{pink}{\texttt{neutral}}} & \multirow{2}{*}{1.80} & \multirow{2}{*}{0.00} & \multirow{2}{*}{1.60}\\
Bool & {\textcolor{premise}P:} Ann or Barbara is optimistic. {\textcolor{hypothesis}H:} Ann is optimistic\\
\midrule
T13 & {\textcolor{premise}P:} \{NAME1\} and \{NAME2\} are from \{CTRY1\} and \{CTRY2\} respectively. {\textcolor{hypothesis}H:} \{NAME1\} is from \{CTRY1\}. {\colorbox{pink}{\texttt{entail}}} & \multirow{2}{*}{57.10} & \multirow{2}{*}{59.20} & \multirow{2}{*}{100.00}\\
Bool & {\textcolor{premise}P:} Mary and David are from Canada and Australia respectively. {\textcolor{hypothesis}H:} Mary is from Canada.\\
\midrule
T14 & {\textcolor{premise}P:} \{NAME1\} and \{NAME2\} are from \{CTRY1\} and \{CTRY2\} respectively. {\textcolor{hypothesis}H:} \{NAME1\} is from \{CTRY2\}. {\colorbox{pink}{\texttt{cont}}} & \multirow{2}{*}{46.00} & \multirow{2}{*}{39.30} & \multirow{2}{*}{49.50}\\
Bool & {\textcolor{premise}P:} Robert and Charles are from Russia and France respectively. {\textcolor{hypothesis}H:} Charles is from Russia.\\
\midrule
T21 & {\textcolor{premise}P:} \{NAME1\} and \{NAME2\} are from \{CTRY1\} and \{CTRY2\} respectively. {\textcolor{hypothesis}H:} \{NAME1\} is not from \{CTRY2\}. {\colorbox{pink}{\texttt{entail}}} & \multirow{2}{*}{0.00} & \multirow{2}{*}{0.00} & \multirow{2}{*}{0.00}\\
Bool & {\textcolor{premise}P:} Margaret and Robert are from America and Russia respectively. {\textcolor{hypothesis}H:} Margaret is not from Russia.\\
\midrule
T22 & {\textcolor{premise}P:} \{NAME1\} and \{NAME2\} are from \{CTRY1\} and \{CTRY2\} respectively. {\textcolor{hypothesis}H:} \{NAME1\} is not from \{CTRY1\}. {\colorbox{pink}{\texttt{cont}}} & \multirow{2}{*}{100.00} & \multirow{2}{*}{100.00} & \multirow{2}{*}{100.00}\\
Bool & {\textcolor{premise}P:} John and James are from India and China respectively. {\textcolor{hypothesis}H:} James is not from China.\\
\midrule
T45 & {\textcolor{premise}P:} \{MALE NAME\} and \{FEMALE NAME\} are \{friends/collegues/married\}. He is \{a/an\} \{PROF1\} and she is \{a/an\} \{PROF2\}. {\textcolor{hypothesis}H:} \{MALE NAME\} is \{a/an\} \{PROF1\}. {\colorbox{pink}{\texttt{entail}}} & \multirow{2}{*}{99.50} & \multirow{2}{*}{98.00} & \multirow{2}{*}{97.70}\\
Coref & {\textcolor{premise}P:} Marlen and Rudolf work together. He is a minister and she is professor. {\textcolor{hypothesis}H:} Rudolf is minister.\\
\midrule
T46 & {\textcolor{premise}P:} \{MALE NAME\} and \{FEMALE NAME\} are \{friends/collegues/married\}. He is \{a/an\} \{PROF1\} and she is \{a/an\} \{PROF2\}. {\textcolor{hypothesis}H:} \{MALE NAME\} is \{a/an\} \{PROF2\}. {\colorbox{pink}{\texttt{cont}}} & \multirow{2}{*}{4.70} & \multirow{2}{*}{2.40} & \multirow{2}{*}{92.60}\\
Coref & {\textcolor{premise}P:} Mujtaba and Teresa are collegues. She is a farmer and he is professor. {\textcolor{hypothesis}H:} Teresa is professor.\\
\midrule
T63 & {\textcolor{premise}P:} \{NAME1\} is \{COM ADJ\} than \{NAME2\} {\textcolor{hypothesis}H:} \{NAME2\} is \{COM ADJ\} than \{NAME1\}. {\colorbox{pink}{\texttt{cont}}} & \multirow{2}{*}{85.30} & \multirow{2}{*}{0.00} & \multirow{2}{*}{100.00}\\
Comp & {\textcolor{premise}P:} Elizabeth is harsher than Caroline. {\textcolor{hypothesis}H:} Caroline is harsher than Elizabeth.\\
\midrule
T68 & {\textcolor{premise}P:} \{NAME1\} is \{COM ADJ\} than \{NAME2\}. \{NAME1\} is \{COM ADJ\} than \{NAME3\}. {\textcolor{hypothesis}H:} \{NAME2/3\} is \{COM ADJ\} than \{NAME3/2\}. {\colorbox{pink}{\texttt{neutral}}} & \multirow{2}{*}{0.00} & \multirow{2}{*}{0.00} & \multirow{2}{*}{0.00}\\
Comp & {\textcolor{premise}P:} Philip is more important than Frances. Philip is more important than Kevin. {\textcolor{hypothesis}H:} Kevin is more important than Frances.\\
\midrule
T71 & {\textcolor{premise}P:} Among \{NAME1\}, \{NAME2\} and \{NAME3\} the \{SUP ADJ\} is \{NAME1\} {\textcolor{hypothesis}H:} \{NAME3/2\} is \{COM ADJ\} than \{NAME2/3\}. {\colorbox{pink}{\texttt{neutral}}} & \multirow{2}{*}{16.80} & \multirow{2}{*}{1.70} & \multirow{2}{*}{0.00}\\
Comp & {\textcolor{premise}P:} Among Dorothy, Peter and Laura the healthiest is Peter. {\textcolor{hypothesis}H:} Dorothy is healthier than Laura.\\
\midrule
T76 & {\textcolor{premise}P:} \{NAME1\} is \{ADJ\}. \{NAME2\} is \{COM ADJ\}. {\textcolor{hypothesis}H:} \{NAME2\} is \{COM ADJ\} than \{NAME1\} {\colorbox{pink}{\texttt{entail}}} & \multirow{2}{*}{56.30} & \multirow{2}{*}{93.70} & \multirow{2}{*}{100.00}\\
Comp & {\textcolor{premise}P:} Emily is poor. Nancy is poorer. {\textcolor{hypothesis}H:} Nancy is poorer than Emily.\\
\midrule
T77 & {\textcolor{premise}P:} \{NAME1\} is \{ADJ\}. \{NAME2\} is \{COM ADJ\}. {\textcolor{hypothesis}H:} \{NAME1\} is \{COM ADJ\} than \{NAME2\} {\colorbox{pink}{\texttt{cont}}} & \multirow{2}{*}{45.90} & \multirow{2}{*}{0.00} & \multirow{2}{*}{88.90}\\
Comp & {\textcolor{premise}P:} Anna is smart. Julia is smarter. {\textcolor{hypothesis}H:} Anna is smarter than Julia.\\
\midrule
T80 & {\textcolor{premise}P:} \{NAME\} was facing \{DIR\} and turned towards his/her \{left/right/back\} {\textcolor{hypothesis}H:} \{NAME\} is now facing \{DIR\} {\colorbox{pink}{\texttt{entail}}} & \multirow{2}{*}{0.00} & \multirow{2}{*}{0.00} & \multirow{2}{*}{0.00}\\
Spatial & {\textcolor{premise}P:} Michael was facing east and turned towards his back. {\textcolor{hypothesis}H:} Michael is now facing west.\\
\midrule
T81 & {\textcolor{premise}P:} \{NAME\} was facing \{DIR\} and turned towards his/her \{left/right/back\} {\textcolor{hypothesis}H:} \{NAME\} is now facing \{DIR\} {\colorbox{pink}{\texttt{cont}}} & \multirow{2}{*}{30.20} & \multirow{2}{*}{30.20} & \multirow{2}{*}{0.40}\\
Spatial & {\textcolor{premise}P:} Jane was facing north and turned towards her back. {\textcolor{hypothesis}H:} Jane is now facing north.\\
\midrule
T82 & {\textcolor{premise}P:} \{CITY1\} is \{N1\} miles from \{CITY2\} and \{N2\} miles from \{CITY3\} {\textcolor{hypothesis}H:} \{CITY1\} is \{nearer/farther\} to \{CITY2\} than \{CITY3\} {\colorbox{pink}{\texttt{entail}}} & \multirow{2}{*}{99.50} & \multirow{2}{*}{82.20} & \multirow{2}{*}{88.90}\\
Spatial & {\textcolor{premise}P:} Hartford is 99 miles from Phoenix and 45 miles from Philadelphia. {\textcolor{hypothesis}H:} Hartford is nearer to Philadelphia than Phoenix.\\
\midrule
T83 & {\textcolor{premise}P:} \{CITY1\} is \{N1\} miles from \{CITY2\} and \{N2\} miles from \{CITY3\} {\textcolor{hypothesis}H:} \{CITY1\} is \{nearer/farther\} to \{CITY2\} than \{CITY3\} {\colorbox{pink}{\texttt{cont}}} & \multirow{2}{*}{0.80} & \multirow{2}{*}{6.70} & \multirow{2}{*}{49.70}\\
Spatial & {\textcolor{premise}P:} San Diego is 82 miles from Sacramento and 27 miles from Boston. {\textcolor{hypothesis}H:} San Diego is nearer to Sacramento than Boston.\\
\midrule
T88 & {\textcolor{premise}P:} \{NAME1\} was born in \{YEAR1\} and \{NAME2\} was born in \{YEAR2\}. {\textcolor{hypothesis}H:} \{NAME1/2\} was born \{before/after\} \{NAME1/2\} {\colorbox{pink}{\texttt{entail}}} & \multirow{2}{*}{2.70} & \multirow{2}{*}{41.50} & \multirow{2}{*}{95.60}\\
Temp & {\textcolor{premise}P:} Grace was born in 2004 and Susan was born in 1998. {\textcolor{hypothesis}H:} Susan was born before Grace.\\
\midrule
T89 & {\textcolor{premise}P:} \{NAME1\} was born in \{YEAR1\} and \{NAME2\} was born in \{YEAR2\}. {\textcolor{hypothesis}H:} \{NAME1/2\} was born \{before/after\} \{NAME1/2\} {\colorbox{pink}{\texttt{cont}}} & \multirow{2}{*}{95.80} & \multirow{2}{*}{35.60} & \multirow{2}{*}{0.00}\\
Temp & {\textcolor{premise}P:} Emma was born in 2016 and Harry was born in 1983. {\textcolor{hypothesis}H:} Harry was born after Emma.\\
\midrule
T92 & {\textcolor{premise}P:} \{NAME1\} was born in \{YEAR1\} and \{NAME2\} was born in \{YEAR2\}. {\textcolor{hypothesis}H:} \{NAME1/2\} was born \{earlier/later\} than \{NAME1/2\} {\colorbox{pink}{\texttt{entail}}} & \multirow{2}{*}{93.00} & \multirow{2}{*}{98.80} & \multirow{2}{*}{99.90}\\
Temp & {\textcolor{premise}P:} Christine was born in 2016 and Marie was born in 1985. {\textcolor{hypothesis}H:} Marie was born earlier than Christine.\\
\midrule
T93 & {\textcolor{premise}P:} \{NAME1\} was born in \{YEAR1\} and \{NAME2\} was born in \{YEAR2\}. {\textcolor{hypothesis}H:} \{NAME1/2\} was born \{earlier/later\} than \{NAME1/2\} {\colorbox{pink}{\texttt{cont}}} & \multirow{2}{*}{0.20} & \multirow{2}{*}{0.10} & \multirow{2}{*}{0.00}\\
Temp & {\textcolor{premise}P:} Mark was born in 2007 and Mike was born in 1987. {\textcolor{hypothesis}H:} Mark was born earlier than Mike.\\
\midrule
T98 & {\textcolor{premise}P:} \{NAME\} has \{EVENT1\} followed by \{EVENT2\}. {\textcolor{hypothesis}H:} \{EVENT1/2\} is \{before/after\} \{EVENT1/2\} {\colorbox{pink}{\texttt{entail}}} & \multirow{2}{*}{50.00} & \multirow{2}{*}{35.40} & \multirow{2}{*}{88.60}\\
Temp & {\textcolor{premise}P:} Karen has lunch followed by a history class. {\textcolor{hypothesis}H:} The history class is after lunch.\\
\midrule
T99 & {\textcolor{premise}P:} \{NAME\} has \{EVENT1\} followed by \{EVENT2\}. {\textcolor{hypothesis}H:} \{EVENT1/2\} is \{before/after\} \{EVENT1/2\} {\colorbox{pink}{\texttt{cont}}} & \multirow{2}{*}{64.20} & \multirow{2}{*}{86.30} & \multirow{2}{*}{96.00}\\
Temp & {\textcolor{premise}P:} Kate has lunch followed by a meeting. {\textcolor{hypothesis}H:} The meeting is before lunch.\\
\midrule
T116 & {\textcolor{premise}P:} \{NAME1\} \{bought/taught/...\} \{OBJ\} to \{NAME2\} {\textcolor{hypothesis}H:} \{NAME2\} \{sold/learnt/...\} \{OBJ\} from \{NAME1\} {\colorbox{pink}{\texttt{entail}}} & \multirow{2}{*}{85.20} & \multirow{2}{*}{100.00} & \multirow{2}{*}{89.40}\\
Causal & {\textcolor{premise}P:} Jennifer taught physics to Rachel. {\textcolor{hypothesis}H:} Rachel learnt physics from Jennifer.\\
\midrule
T117 & {\textcolor{premise}P:} \{NAME1\} \{bought/taught/...\} \{OBJ\} to \{NAME2\} {\textcolor{hypothesis}H:} \{NAME1\} \{sold/learnt/...\} \{OBJ\} from \{NAME2\} {\colorbox{pink}{\texttt{cont}}} & \multirow{2}{*}{24.00} & \multirow{2}{*}{1.70} & \multirow{2}{*}{71.50}\\
Causal & {\textcolor{premise}P:} Barbara lent money to Laura. {\textcolor{hypothesis}H:} Barbara borrowed money from Laura.\\
\midrule
T122 & {\textcolor{premise}P:} None the \{OBJS\} are \{COLOR\} in color. {\textcolor{hypothesis}H:} Some of the \{OBJS\} are \{COLOR\} in color. {\colorbox{pink}{\texttt{cont}}} & \multirow{2}{*}{0.00} & \multirow{2}{*}{0.00} & \multirow{2}{*}{100.00}\\
Quan & {\textcolor{premise}P:} None of the wallpapers are green in colour. {\textcolor{hypothesis}H:} Some of the wallpapers are green in colour.\\
\midrule
T171 & {\textcolor{premise}P:} \{OBJ1\} and \{OBJ2\} lie on the table. \{NAME\} asked for \{OBJ1\} {\textcolor{hypothesis}H:} \{NAME\} also asked for \{OBJ2\} {\colorbox{pink}{\texttt{cont}}} & \multirow{2}{*}{0.00} & \multirow{2}{*}{0.00} & \multirow{2}{*}{0.00}\\
Impl & {\textcolor{premise}P:} Silverware and plate lie on the table. Barbara asked for the plate. {\textcolor{hypothesis}H:} Barbara also asked for the silverware.\\
\midrule
T172 & {\textcolor{premise}P:} \{OBJ1\} and \{OBJ2\} lie on the table. \{NAME\} asked for \{OBJ1\} {\textcolor{hypothesis}H:} \{NAME\} also asked for \{OBJ2\} {\colorbox{pink}{\texttt{neutral}}} & \multirow{2}{*}{33.80} & \multirow{2}{*}{18.40} & \multirow{2}{*}{99.80}\\
Bool & {\textcolor{premise}P:} Pin and pitcher lie on the table. Peter asked for the pitcher. {\textcolor{hypothesis}H:} Peter also asked for the pin.\\
\midrule
T128 & {\textcolor{premise}P:} Some the \{OBJS\} are \{COLOR\} in color. {\textcolor{hypothesis}H:} All of the \{OBJS\} are \{COLOR\} in color. {\colorbox{pink}{\texttt{cont}}} & \multirow{2}{*}{100.00} & \multirow{2}{*}{100.00} & \multirow{2}{*}{100.00}\\
Impl & {\textcolor{premise}P:} Some of the balls are purple in colour. {\textcolor{hypothesis}H:} All of the balls are purple in colour.\\
\midrule
T127 & {\textcolor{premise}P:} Some the \{OBJS\} are \{COLOR\} in color. {\textcolor{hypothesis}H:} All of the \{OBJS\} are \{COLOR\} in color. {\colorbox{pink}{\texttt{neutral}}} & \multirow{2}{*}{0.00} & \multirow{2}{*}{0.00} & \multirow{2}{*}{0.00}\\
Quan & {\textcolor{premise}P:} Some of the cars are green in colour. {\textcolor{hypothesis}H:} All of the cars are green in colour.\\
\arrayrulecolor{black}\bottomrule
\end{tabular} 
}
\caption{We show some interesting Templates, and model accuracies on the templates.}
\label{tab:app:interestingtemplates}
\end{table*}

\end{document}